\documentclass{scrartcl}
\usepackage[maxbibnames=20,style=authoryear]{biblatex}
\usepackage{microtype}
\usepackage{booktabs}
\usepackage{graphicx}
\usepackage[T1]{fontenc}
\usepackage{lmodern}
\usepackage{hyperref}
\usepackage{colortbl}

\title{Towards Computational Analysis of Gender Depiction in the Comedias of Calderón de la Barca}
\author{Allison Keith \and Antonio Rojas Castro \and Hanno Ehrlicher \and Kerstin Jung \and Sebastian Padó}
\date{}

\newcommand{\al}[1]{}
\newcommand{\an}[1]{}
\newcommand{\ha}[1]{}
\newcommand{\ke}[1]{}
\newcommand{\se}[1]{}

\begin{document}

\maketitle

\begin{abstract}
In theater, playwrights can use the portrayal of characters to explore culturally based gender norms. In this paper, we develop quantitative methods to study gender depiction in the non-religious works (\textit{comedias}) of Pedro Calderón de la Barca, a prolific Spanish 17th century author. We gather insights from a corpus of more than 100 plays by using a gender classifier and applying model explainability (attribution) methods to determine which text features are most influential in the model’s decision to classify speech as `male’ or `female’, indicating the most gendered elements of dialogue in Calderón's \textit{comedias} in a manner accessible to humans. We find that the classifier picks up the male/female distinction mostly successfully (up to f=0.83), yields semantic aspects of gender portrayal, and demonstrates that the model is useful in providing a relatively accurate scene-by-scene prediction of cross-dressing characters.
\end{abstract}

\section{Introduction}

In Spain, the Baroque period (approx. 1600–1700) was a period of immense artistic creativity, generally known as the `Golden Age’ (\textit{Siglo de Oro}). This is particularly true in literature, where the period saw exceptional writers such as Lope de Vega, Tirso de Molina and Pedro Calderón de la Barca. The latter, who lived from 1600 to 1681, is generally considered as one of the most important playwrights of the age. He was immensely productive, writing a total of over 200 theatrical plays, both secular and religious, which had a lasting impact on Spanish theatre and beyond (\cite{fitzmaurice_1921}). He is particularly known for detailed and nuanced portrayal of the characters in his works (\cite{woods_2013}).

Not surprisingly, Calderón's writings have been subject to intense analysis by literary scholars over a long period of time, and topics have moved in and out of fashion. For example, traditional foci of scholarship have been the role of honor and power (\cite{greer_2017}) as well as his attention to dramatic structure (\cite{whitby_1954}).

A relatively new aspect among these, which has gained global attention in Hispanic Studies since the latter half of the 20th century, is \textit{gender depiction}, that is, the question of how Calderón conceptualized male and female roles in his plays differently (\cite{arellano_2015,saez_2019,tietz_2017}). Interestingly, scholars have come to very different conclusions regarding Calderón's gender portrayal. Some studies find that Calderón depicts a broadly period-typical understanding of gender identity and gender norms. For example, \textcite{ibarreche_2020} study the representation of male characters, particularly in relation to honor, in plays such as \textit{El médico de su honra}, \textit{El Alcalde Zalamea}, or \textit{El pintor de su deshonra}. In these pieces, male characters, often husbands, fathers, or brothers, strive to uphold their social status, wielding power, owning properties, and navigating public spaces with traits such as bravery, patience, and caution (\cite{fernandez_2017}). Conversely, in plays like \textit{Casa con dos puertas mala es de guardar} or \textit{La dama duende}, women are frequently confined to domestic spaces, initiating disputes and deftly manipulating characters like puppets in games of seduction leading to a harmonious resolution (\cite{saez_2019}).

In contrast, other studies have found that Calderón's work transcends societal norms by providing a dynamic platform for the exploration of gender identities (\cite{tietz_2017}) such as ``perverse’’ women -- women who use deception against men to achieve their goals. While absent from comedies, these form a recurring character type in tragic pieces (\cite{arellano_2015}). Similarly, several well-known works depict female characters cross-dressing as men and performing traits typically associated with masculinity (\cite{bravo_1976, strosetzki_2017, araico_2017}). A commonly used trope of the time period, playwrights utilized cross-dressing as a means for female characters to transcend the domestic space and actively participate in both society and the action of the play. 

Thus, there appear to be conflicting tendencies of gender depiction in Calderón's work: It is both deeply shaped by plot and genre conventions, but is used as a grounds to expand social ideas about gender roles by representing women with agency who drive plot progression and utilize covert ways to exercise their will (\cite{woods_2013}). A possible explanation for the picture that emerges is that, throughout the massive catalogue of Calderón's works, a select few have been given close attention by scholars who adopt a `close reading’ method focusing on individual dramas sampled of the more than 100 secular dramas written by Calderón. Consequently, studying all the dramas can show if this picture holds true for the oeuvre as a whole.

This prompts the classical motivation for scalable reading methods (\cite{weitin_2017}) from digital humanities, combining computational models with manual interpretation: Can we make out \textit{global} patterns in gender depiction that hold across Calderón's whole body of secular dramas?

In this study, we propose to build on recent developments regarding the \textit{explainability of AI models} (\cite{wiegreffe_2019,bibal_2022}) which aim at providing a human-understandable justification of model decisions. This makes it possible to consider a computational model of a literary phenomenon primarily as an \textit{aggregation mechanism} that summarizes large amounts of text into a much smaller set of predictions, and then to trace these predictions back to the linguistic properties that gave rise to them.

We show the feasibility of this research program by training a \textit{gender classifier} based on the characters' speech. The motivation here is not to carry out the task of gender classification, but to ask questions about the classifier's predictions: How difficult is the task -- i.e., how consistent is Calderón's gender depiction? What features does the model rely on for its predictions -- i.e., what linguistic patterns does Calderón use for gender depiction? What characters or character groups are easy or difficult to classify -- i.e., what characters are depicted in unconventional or unexpected fashions?

Our work is not the first to classify characters and their properties based on their speech. Five are particularly close. First, \textcite{vishnubhotla_2019} analyze the ability of classifiers to distinguish among characters within a play to determine if authors were able to write characters with distinct voices; they however ask the question at a general level, without reference to gender or Calderón. Second, \textcite{bullard_2014}  predict various demographic properties of characters, including gender. However, their study was still based on hand-picked features fed into a regression model. Next, \textcite{savoy_2023}  conducted a classification experiment on the gender of Shakespeare characters based on lines, using random forest and logistic regression. Two more studies adopt tools from stylometry, extending the approach from authorship attribution to the more fine-grained character portrayal.  \textcite{vsecla_2024} carry out $n$-gram analysis to examine characteristics of gender depiction in French, Russian, German, and English dramas, specifically looking at the distinctiveness of characters and revealing distinct authorial style. Finally, \textcite{lorenzo_2024} compares the depiction of male and female characters in the works of Calderón and another playwright, Sor Juana de la Cruz, on a comparatively small number of works.

Our study takes advantage of recent developments in Natural Language Processing by representing character speech using language models (LMs). This  gives rise to two major new design questions. First, LMs decrease reliance on feature engineering, but for the purpose we have in mind (explaining the model predictions), not all parts of the input are equally informative. Second, LMs are still known to struggle with identifying relevant information in long texts. This raises the question of how much information to present to the classifier at each time: Too little (e.g., just one line), and the input may contain insufficient hints as to the speaker's gender; too much (e.g., a character's complete text from a drama), and the model may struggle. Also, the predictions of the model for each individual part of the input have to be aggregated into a prediction at a more global level, for which we also consider multiple options. We discuss both of these issues in \autoref{sec:methods} below.

We find that our models are able to correctly classify the gender of characters in Calderón's \textit{comedias}. Our experiments yield the best results when given the most amount of input text, but the models also perform well when using an aggregation method that takes into account the confidence of each prediction. We take this as a positive result for computational analysis of gender portrayal in the works of Calderón: It shows that there are recognizable gender cues at the scene level and, at the same time, that characters' gender roles can vary over the course of a play. Finally, an examination of the most explanatory features in the models predictions demonstrate that the model picks up on both grammatical and semantic information that contribute to gender prediction. The examination of cross-dressing female characters demonstrates that these characters' lines are more similar to the speech of male characters compared to more traditional female characters, particularly during cross-dressing scenes. 

\section{Methods}
\label{sec:methods}

\subsection{Language Models and Word Embedding-Based Classification}

The most important direction of innovation in natural language processing (NLP) of the last ten years has been the development of \textit{distributional semantics} into a framework which underlies almost all processing models in NLP. This idea reached its current form through four major phases.

In the first phase, distributional semantics, built on ideas from structuralist semantics (\cite{firth_1957}), was developed into an approach that aimed at capturing the meaning of target words by counting co-occurrences of these target words with words in their context. This naturally corresponds to a representation of words as context vectors (or \textit{embeddings}), and of semantic similarity as a measure of contextual similarity between words (\cite{miller_1991}). 

In the second phase, the construction of such embeddings moved from explicit context counting to the use of neural language models. Such language models are presented with a textual prefix (`holiday on Sri \dots’) and have the task of predicting the subsequent token (`Lanka’). When trained to do so on large text corpora, the learned parameters of such language models are essentially word embeddings (\cite{mikolov_2013}). 

The next phase involved the introduction of the so-called transformer architecture, which introduces the ability to make word embeddings sensitive to their context: e.g., `root’ receives a different embedding in the phrase `root canal’ than in the phrase `square root’. The resulting \textit{contextualized} language models can account for lexical ambiguity in a principled fashion (\cite{devlin_2019}). Model families of this type, such as BERT, have shown to be applicable to almost all standard tasks in NLP by obtaining word embeddings for an input text, aggregating these embeddings to obtain a text embedding (e.g. simply by averaging), and passing the aggregate through a very simple decision mechanism -- e.g., a single-layer neural network with two output units which directly corresponds to a logistic regression model over the text embedding. Such models are typically trained in two phases: First the language model is `pre-trained’ on a large amount of unlabeled text, then the decision layer is `fine-tuned’ on a comparatively small amount of task-specific labeled text.

The final phase in the development of language models consists in the emergence of large language models (LLMs) which result from massive scaling up of the model training. The resulting LLMs are to a substantial degree task-agnostic (i.e., do not require fine-tuning) and can be instructed (`prompted’) directly to carry out some task (\cite{liu_2023a}). To solve a classification task, for example, models can be asked to answer `yes’ or `no’ similar to the way a human annotator would be instructed.

For the task we consider in this study -- predicting the gender of a character in a drama based on (some part of) their utterances -- both contextualized embedding models and LLMs are plausible candidates to develop classifiers. We decided to build our classifier on top of a contextualized embedding model. Such models are typically much smaller (and thus computation efficient), they are more transparent (the details of the training processes of almost all LLMs are confidential (\cite{liesenfeld_2023})), and prompt-based approaches face problems regarding robustness and bias (\cite{webson_2022}).

\subsection{Design Decisions: Input Granularity and Aggregation} \label{design descisions}

As outlined above, a central question that must be decided when presenting text to a LM for classification is the length of the input for which the model is supposed to make a single prediction, that is, the \textit{input granularity}. Short inputs may consist of utterances which are not informative regarding the speaker's gender. As for long inputs, older LM-based models have generally a limited input size (often 512 tokens)\footnote{See \autoref{Experimental Setup} for details about tokenization.} and while newer LLM-based models can deal with much longer inputs, they struggle with identifying the relevant information, in particular if it is in the middle of the text (\cite{liu_2023b}).

\begin{figure}
  \centering
  \includegraphics[width=0.75\linewidth]{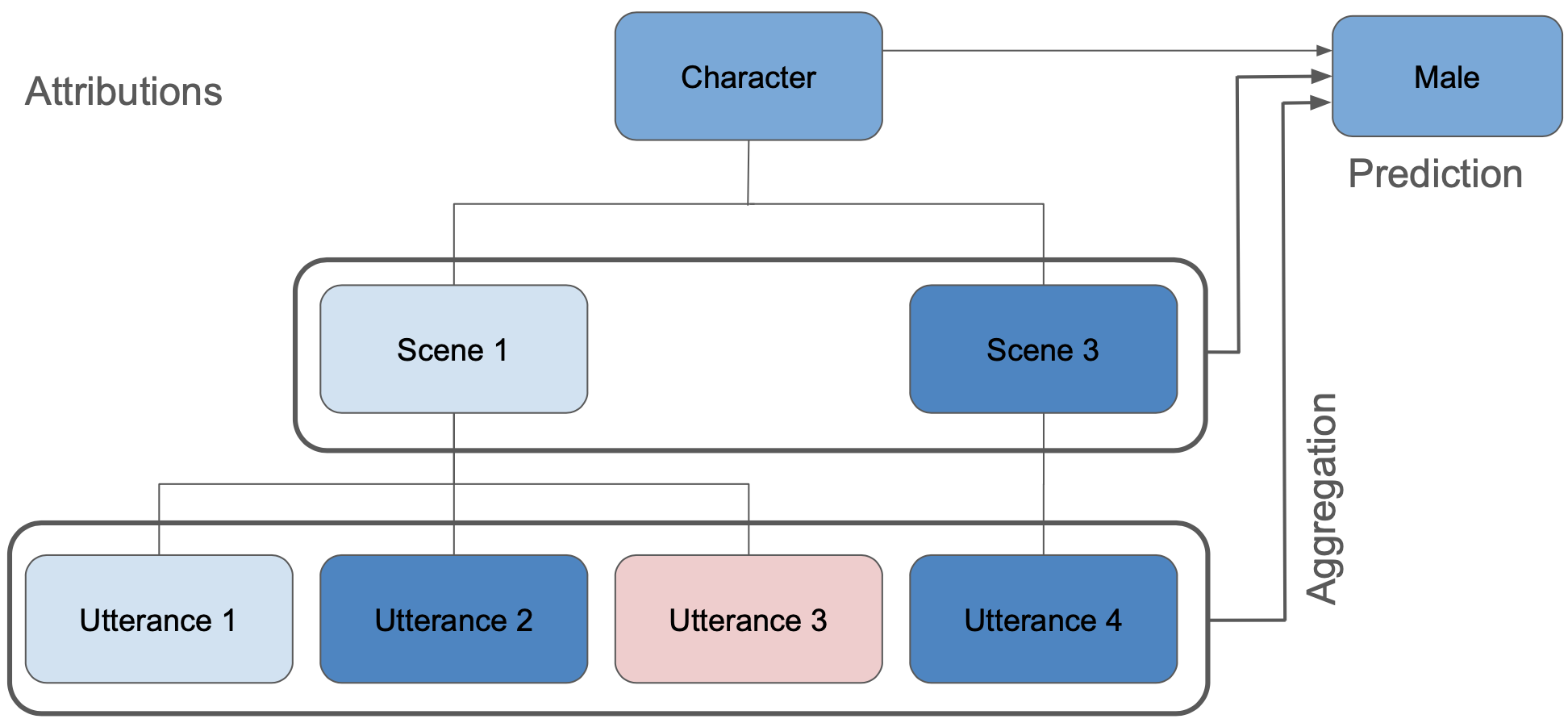}
  \caption{Gender classification for a character at different levels of granularity (blue = masculine, red = feminine). Saturation indicates model confidence.} 
  \label{fig:levels}
\end{figure}

\autoref{fig:levels} shows a simplified view of the hierarchical structure of a character's utterances in a drama: At the bottom we have individual utterances, which vary in length from single sentences to long monologues. Utterances can be grouped together at the scene level, and then again at the global character level. Models can be trained to make gender predictions at each of these levels (shown in blue and red, respectively), and predictions can vary in confidence (shown by saturation). Since we do not know a priori which of these levels works best, our experiment will evaluate models that have been trained at each of these three levels. 

If we choose to classify at a more fine-grained level (such as utterances), there is no guarantee that all utterances of one character are classified in the same way -- in fact, given our assumption that at least some utterances are uninformative regarding gender, complete consistency would be surprising (utterances 2 vs. 3 in \autoref{fig:levels}). Therefore, we also need to define an aggregation method to return a character-level response based on the concrete utterance- or scene-level predictions.\footnote{Utterance-level predictions could also be aggregated at the scene level. We do not consider this option to simplify the experimental setup.} We consider two aggregation strategies. The simple one is \textit{majority} vote: We count the number of utterances (or scenes) predicted as male or female, respectively, and assign the more frequent gender. In \autoref{fig:levels}, 3 out of 4 utterances are classified as male, which leads to an overall classification of the character as male. The second strategy makes use of the fact that LM-based models do not just output labels, but also probabilities. Assuming that more informative inputs are classified more confidently, taking these probabilities into account should give more weight to informative utterances. Formally, we use the \textit{geometric mean} $p_{GM}(g|c)$ of a gender $g$ for a character $c$ given the individual utterances $u(c)$ for the character: 

\begin{equation}
  p_{GM}(g|c) = \left( \prod_{u(c) \in U(c)}^{x} p(g|i_c) \right) ^{\frac{1}{x}}
\end{equation}

In \autoref{fig:levels}, utterance 1 is  low probability male, utterances 2 and 4 are high probability male, and utterance 3 is low probability female, leading to a character-level aggregated classification of male.

\begin{figure}
  \centering
  \includegraphics[width=0.75\linewidth]{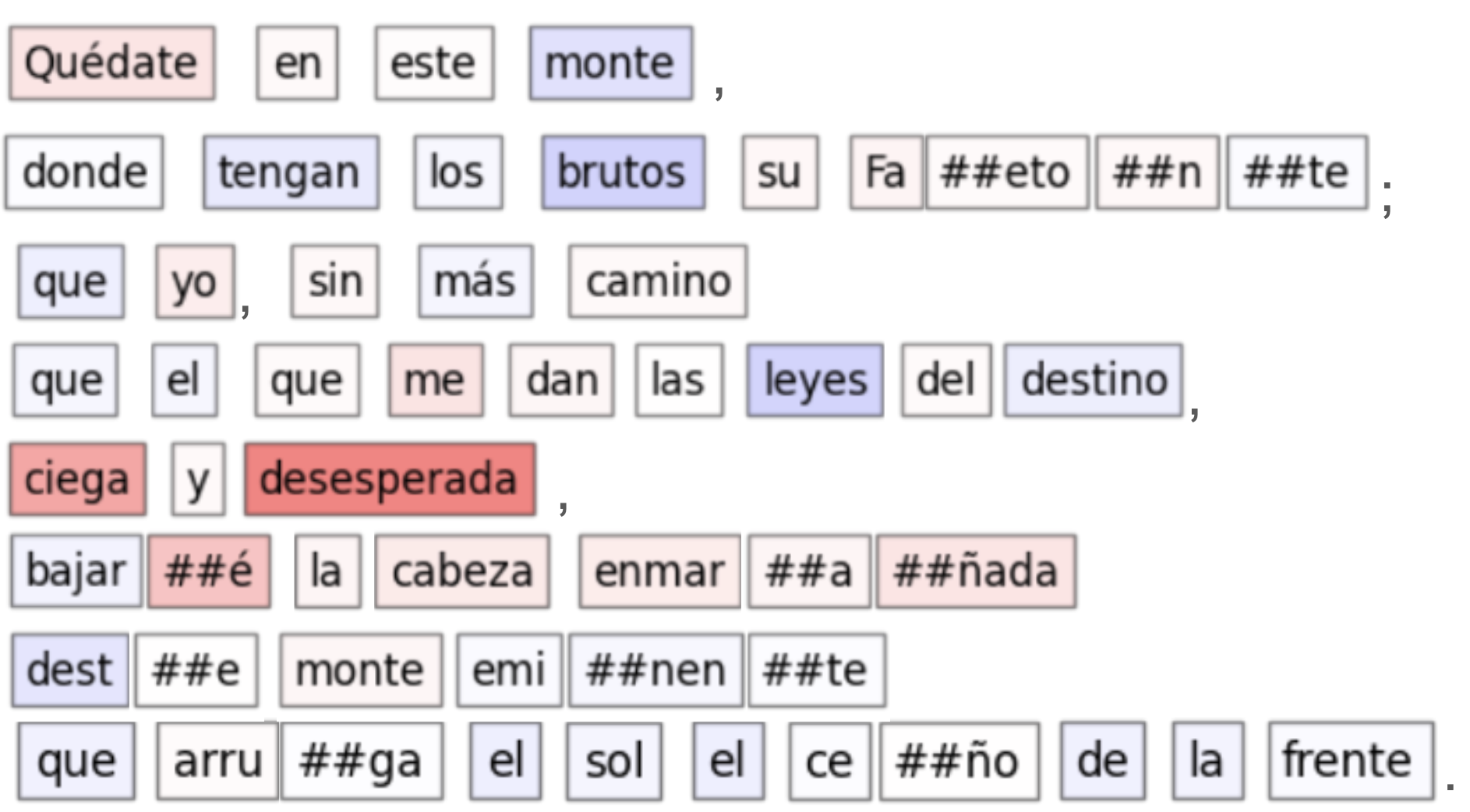}
  \caption{Visualization of the attribution model: blue indicates male, orange female; saturation corresponds to strength of cue. Original passage (\textit{La vida es sue\~no}, Rosaura, Act 1 Scene 1): \textit{Stay on this mountain, where the brutes have their Phaeton, and I, with no other path than the one given to me by the laws of destiny, blind and desperate, will lower my tangled head from this eminent mountain where the sun wrinkles the frown on my forehead.}}
  \label{fig:enter-label}
\end{figure}

\subsection{Analyzing Neural Classifiers with Attribution Methods} 

The topic of making machine learning models  explainable has seen considerable attention in recent years, from a variety of motivations (\cite{riedl_2019}). Two major families (\cite{murdoch_2019}) are model-based approaches -- which build on the inherent interpretability of small models -- and post-hoc methods -- which analyze previously trained models to explain their predictions. In turn, post-hoc methods can be grouped into global and local methods. Global methods analyze the entirety of a model's predictions over some dataset, while local models explain individual predictions (\cite{atanasova_2020}). In our study, we focus on local methods because we are interested in looking at individual predictions. A prominent local approach for the attention-based models ubiquitous in NLP is the analysis of attention weights; however, it has been shown that these models are not guaranteed to faithfully attribute the model’s prediction to its input features (\cite{jain_2019}). Therefore, we adopt the so-called \textit{integrated gradient} approach by \textcite{sundararajan_2017}. Applied to a binary classification task with classes A and B, this approach explains every prediction as a sum of weights (positive, i.e, supporting class A, or negative, i.e., supporting class B) over the input features.  In our case, every word in the input text is assigned a weight that indicates how strongly it supports (or contradicts) the assignment of a particular gender to the text. 

\autoref{fig:enter-label} shows some example sentences from a Calderón drama, where orange color indicates words that are typical for female speech and blue color for male speech. Again, saturation corresponds to the strength of the cue. Note that these attributions can combine morphological and semantic considerations: The most strongly female token, ``desesperada’’ (\textit{desperate}), is more likely to be uttered by a woman both due to its female suffix \textit{-a}, and by topical association.

\subsection{Lexical Overfitting} 

A final complication in our use of explanation methods to mine a classifier for information about an author's linguistic patterns is that the features presented to the classifier need to be \textit{informative} for the research question. This is not guaranteed, since a classifier can base its decision on features that are uninteresting from a theoretical perspective. For example, person names are informative for gender classification since men and women interact more within their genders than across. At the same time, person names are  arguably uninteresting with respect to gender portrayal. In Calderón's tragedy \textit{Un castigo en tres venganzas}, for example, only the main character Federico ever addresses the character Becoquín -- meaning when the token \textit{Becoquín} appears, it is with certainty spoken by the character Federico and thus a perfect cue for male gender. More generally, the problem is the Zipf distribution of natural language, as a consequence of which there is a large number of infrequent words that occur either only in male or in female speech and could therefore be identified by the classifier as highly informative for the respective class. 

This can be seen as an instance of overfitting (\cite{wolfe_2021}). The problem is also well known in stylometry (\cite{evert_2017,vsecla_2024}). It is usually addressed by requiring words to meet certain frequency conditions, such as top-$n$ words,  mid-frequent words (\cite{caliskan_2022}) or words that occur in a minimum number of documents. Another strategy is to mask particularly predictive tokens (\cite{tang_2022}), thus making them unavailable to the classifier and forcing it to focus on more general cues.

\section{Experimental Setup}\label{Experimental Setup}

To recapitulate, we structure our study into two experiments. In the first experiment (\autoref{sec:exp1-results}), we aim to see if a language model can generally predict the gender of a character given different amounts of speech and different attribution methods. This experiments adopts a two-factor design. The first factor we vary is the input length, and the second one is the aggregation strategy.

The three input levels are 1) character level, in which we consider all the lines that a character speaks throughout the course of a play as one data-point, 2) scene level, in which we consider all of these lines, but divided by the scene barriers so that each scene is predicted separately, and 3) utterance level, in which we consider only unique lines (or utterances) spoken by a character. The aggregation strategies are: 1) none, 2) majority label, and 3) geometric mean, as discussed in \autoref{sec:methods}. This experiment targets the maximum number of characters from Calderón's plays for whom we can obtain reliable digitized text (see below for details). 

The second experiment (\autoref{sec:exp2-results}) focuses specifically on cross-dressing characters, who exhibit a broad range of social roles in Calderóns plays. In his own words, spoken by Rosaura in \textit{La vida es sueño}: 

\begin{quote}
Tres veces son las que ya / me admiras, tres las que ignoras / quién soy, pues las tres me has visto / en diverso traje y forma. / La primera me creíste / varón, en la rigurosa / prisión, donde fue tu vida / de mis desdichas lisonja./ La segunda me admiraste / mujer, cuando fue la pompa / de tu majestad un sueño,/ una fantasma, una sombra./ La tercera es hoy, que siendo / monstruo de una especie y otra,/ entre galas de mujer / armas de varón me adornan.

Three times you have / admired me, three times you do not know / who I am, for three times you have seen me / in different costume and form./ The first time you believed me to be / a man, in the rigorous / prison, where your life was / a flattery of my misfortunes./ The second you admired me / woman, when the pomp / of your majesty was a dream,/ a phantom, a shadow./ The third is today, that being/ a monster of one kind and another,/ among the finery of a woman,/ the weapons of a man adorn me.
\end{quote}

Investigating specifically these characters tells us to what extent the model potentially only picks up `prototypical’ scenes of characters, and to what extent it is able to pick up more fine-grained change of roles across scenes within characters. 

\paragraph{Data.} We base our study on all 109 of Calderón's plays that have previously been classified as non-religious plays of various genres, jointly known as `\textit{comedias}’, which are available in the CalDraCor part of the DraCor project (\cite{fischer_2020}).\footnote{See: \url{https://dracor.org/cal}.} These are orthographically modernized and normalized digital versions of the plays in XML-TEI format. DraCor segments texts by utterances in the style of classical play scripts, with lines labeled with the speaker. The plays are already annotated by act number but lack formal scene division. We enrich all plays with scene divisions, determined by stage directions that indicate a character leaving or entering the stage, to obtain a more fine-grained segmentation. 

In the cast list of each play, all characters are annotated with role information, including `male’ and `female’ labels. These annotations are based on examination of the texts done by literary scholars as part of the DraCor project. Note that the intention of this paper is not to perpetuate the idea of a gender binary, nor to make assertions about gender identity. 

In order to define a simple binary classification task, we leave out a small number of characters who do not have a defined gender in the corpus, typically collectives (such as ``musicos’’ or ``coro’’ (\textit{musicians, choir}). After also removing minor characters who speak less than 30 words -- presumably not enough for reliable classification -- we are left with a total of 1,515 characters (1,021 = Male, 494 = Female): We have roughly a 2:1 male/female character ratio. Males speak a mean of 1,219 tokens (min = 31, max = 7,166) and females on average 1,130 tokens (min = 32, max = 5,388). On this basis, we assume that text length is not a major confounding variable for the male/female distinction. 

We create an 80\%/10\%/10\% train/test/validation split at the character level, ensuring that each character occurs only in one of the partitions (1,207/152/152 characters). Since we only provide the character speech to the classifier, this means that the test and validation partitions are unseen.

\paragraph{Preprocessing.} To avoid the lexical overfitting problem discussed above, we mask all names of characters that appear in the plays (751 unique names)  with the token `[NAME]’. We also mask all tokens that appeared in only one play with a `[MASK]’ token. We also considered removing signals of grammatical gender such as self-predications of the speaker with gender-specific nouns or gendered suffixes (such as ``desesperada’’ above). However, automating such a filter would require substantial linguistic analysis to recognize which predications apply to the speaker themselves, arguably a non-trivial process that runs the risk of introducing inconsistencies. Also, these cases appear sufficiently rarely in the corpus to not have a major effect, as we demonstrate below.

\paragraph{Model Implementation.} Following the argument from the previous section, we build our classifiers by fine-tuning a pre-trained Language Model on our drama data. We tested the model performance on a series of multilingual BERT, DistilBERT, and RoBERTa models. We found the best performance for the classification task was achieved when using BETO (\cite{canete_2020}), a BERT base model pre-trained on Spanish data, namely 3B token from Spanish Wiki, Europarl, and other corpora. We adopt this LM for our experiments. It provides a suitable basis given that the dramas we work with are modernized and normalized.

Note that BETO, like all BERT-style models, carries out internal tokenization using a variant of byte pair encoding (\cite{sennrich_2016}). This method has the goal of limiting the vocabulary of tokens in text. Its consequence is that frequent words tend to remain separate tokens, while infrequent words are split into subword units, minimally individual characters. By convention, subword tokens that start inside original words are prefixed with `\#\#’ (\autoref{fig:enter-label}). Also, BETO has a maximum context window size of 512 tokens, meaning that for longer texts, only the first 512 tokens are taken into account in any processing. This amount of text is generally sufficient for the model to make predictions. Details on model training can be found in  \autoref{sec:training-details}.

\paragraph{Evaluation Design and Metrics.}
\label{Model Evaluation}
We evaluate both the individual predictions and the aggregated predictions for characters quantitatively with precision, recall, and F1 score, macro-averaged between the `male’ and `female’ classes. We use macro averaging because, while the male class is over-represented compared to the female class, we are equally interested in male and female predictions. 

For the qualitative evaluation, we use a library that provides an implementation of the integrated gradients approach (Transformers Interpret Sequence Classification Explainer).\footnote{See: \url{https://github.com/cdpierse/transformers-interpret}.} It yields the contribution of each token towards the overall class prediction of each test input sequence. 

\paragraph{Baseline.} As a baseline model, we compare our LM to the case of always assigning the most frequent class (male). This baseline can be applied at the different input levels. For example, at the character level, where we have 76\% male characters, it achieves a 0.76 F1 score for male characters (R=1, P=0.67) but a 0 F1 score on female characters (R=0, P=undef.). This results in a macro F1 of 0.38.

\section{Results}
\subsection{Experiment 1: General Gender Classification}
\label{sec:exp1-results}

\paragraph{Impact of Input Levels.} \autoref{tab:model_performace} shows the main results for gender classification at the three input levels without any aggregation. The model performs best at the character level (F1 = 0.83), followed by the scene (F1 = 0.70) and utterance (F1 = 0.68) levels. This is not very surprising, given that the character level provides the model with the longest, and thus most informative, input texts. It is interesting, though, that the performance of the model drops by 15 points F1 score between the character and the utterance level while there is hardly any difference between the scene level and the utterance level. The baseline results change hardly between levels, indicating that this effect does not reflect a change in the difficulty of the task. We therefore draw as our first conclusion that reliable gender identification requires more information than is typically available from a single utterance or even scene.

\begin{table}
    \begin{tabular}{lcccccccc}
    \toprule
& \multicolumn{3}{c}{Baseline}  & \multicolumn{3}{c}{Model Results}  

       \\ \\
        Input level $\downarrow$ & Precision & Recall & F1  & Precision & Recall & F1 \\ 
        \cmidrule(r){2-4}
        \cmidrule(lr){5-7}
        \cmidrule(r){2-4}
        \cmidrule(lr){5-7}
        Character per drama & 0.33 & 0.50 & 0.38 &\textbf{0.83} & \textbf{0.82}  & \textbf{0.83} \\
        Character per scene   & 0.33 & 0.50 & 0.40 &0.75 & 0.69 & 0.70  \\ 
        Character utterance & 0.34 & 0.50 & 0.41 & 0.71 & 0.67 & 0.68 \\ 
 \bottomrule
    \end{tabular}
    \caption{Experiment 1: Macro-averaged precision, recall and F1 for gender at different input levels, evaluated  on individual predictions. Baseline = most frequent class. Best results for each measure in boldface.}
    \label{tab:model_performace}
\end{table}

\begin{table}
\begin{tabular}{lcccc}
\toprule
    & Q1 & Q2 & Q3 & Q4 \\
    \cmidrule{2-5}
    length & 4-20 & 21-50 & 51-121 & 122-512 \\
 F1 for character per scene & 0.62 & 0.64 & 0.73 & 0.82 \\
\cmidrule{2-5}
length & 2-9 & 10-14 & 14-26 & 27-512 \\
F1 for character utterance & 0.63 & 0.65 & 0.69 & 0.78 \\
\bottomrule
\end{tabular}
    \caption{F1 for test set characters divided into quartiles by number of words spoken.}
     \label{tab:accuracyXwords}
\end{table}

To explore the failure modes, \autoref{tab:accuracyXwords} shows F1 scores for our gender classifiers at the scene and utterance levels. We split these numbers  into four quartiles of characters in terms of numbers of words spoken. The results again demonstrate a clear improvement for longer texts of 15-20 points F1 score between the first and the forth quartile. Evidently, many short utterances are not informative enough to permit reliable gender determination of the speaker. 

\paragraph{Impact of Aggregation Method.} 
Next, \autoref{tab:model_performace_aggregation} presents the performance of the models when classification is first carried out at the utterance or scene level and then aggregated to the character level. The numbers in \autoref{tab:model_performace_aggregation} are comparable to the `Character per drama’ row of \autoref{tab:model_performace}. 

\begin{table}
    \begin{tabular}{lcccccccc}
    \toprule
Aggregation $\rightarrow$ &  
       \multicolumn{3}{c}{Majority label} & 
       \multicolumn{3}{c}{Geometric mean} & 
       \\ \\
        Input level $\downarrow$ & Precision & Recall & F1  & Precision & Recall & F1  \\ 
        \cmidrule(r){2-4}
        \cmidrule(lr){5-7}
        \cmidrule(r){2-4}
        \cmidrule(lr){5-7}
        Character per scene  & 0.51 & 0.51 & 0.49 & \textbf{0.81} & \textbf{0.81} & \textbf{0.81} \\ 
        Character utterance & 0.33 & 0.31 & 0.32 & 0.75 & 0.79 & 0.76 \\ 
 \bottomrule
    \end{tabular}
    \caption{Experiment 1: Model performance when gender is predicted at the utterance or scene level and then aggregated to the character level, either with majority vote or with geometric mean.}   
    \label{tab:model_performace_aggregation}
\end{table}

Comparing the two aggregation methods, we observe that the majority vote performs rather badly, leading to a substantial loss in performance compared to the predictions without aggregation in \autoref{tab:model_performace}. In contrast, the combination by geometric mean performs well, and aggregated predictions at the scene level almost reach the performance of the model that considers the complete text of a character.

This pattern meshes well with our interpretation above: When we use the probability estimates of the model, uninformative utterances have less impact on the overall prediction, and performance improves. All in all, we believe that the \textit{aggregated scene-level model} is the most interesting among the ones we consider, since it is almost as good as the character-level model when predictions at the character level are evaluated, but also provides a separate prediction for each character in each scene, making it possible to pick up changes in gender roles over the course of a play -- exactly the phenomenon that we are going to investigate below in Experiment~2 (\autoref{sec:exp2-results}). Thus, our second analysis yields a somewhat more nuanced picture: Gender classification requires a fair amount of text, but the effect can be mitigated by informed aggregation.

\paragraph{Predictions for Male vs. Female Characters.} 

\begin{table}
    \begin{tabular}{lcccccc}
    \hline
         Aggregation $\rightarrow$ & \multicolumn{2}{c}{None} & \multicolumn{2}{c}{Majority Label} & \multicolumn{2}{c}{Geometric Mean}   \\
         Input Level $\downarrow$ & Male & Female & Male & Female & Male & Female \\
         Character per drama & \textbf{0.87} & 0.70 & - & - & - & - \\
         Character per scene & 0.83 & 0.58 & \textbf{0.72} & 0.25 & \textbf{0.87} & 0.74 \\
         Character utterance & 0.82 & 0.54 & 0.51 & 0.12 & 0.82 & 0.71 \\
    \hline
    \end{tabular}
    \caption{Model F1-Scores comparing performance on predictions of male and female characters.}
\label{tab:results_by_gender}
\end{table}

\autoref{tab:results_by_gender} shows the results of the non-baseline models broken down by gender; best results for each metric and each gender are boldfaced. The results for male characters are generally better than for female characters, which is expected, given the higher number of male characters. That said, there is a striking difference in the effect of the aggregation strategy between genders: While results for male characters show only a rather small change of 5 points between the character and utterance levels, results for female drastically improve at the higher input levels (+.4 at the scene level and +16 at the drama level). When taking into account just the majority label for each character, the performance drops, especially for the female characters, at both the scene and utterance levels. When calculating the geometric mean, the performance at the scene and utterance level for both male and female characters increases again, in particular for female characters. Again, this goes well with our general observation that some sentences are highly gendered while others are uninformative as far as the model is concerned. Thus, the geometric mean-based aggregation can be seen as a strategy to reduce gender bias, according to its definition as performance difference between genders (\cite{rudinger_2018}).

To further validate this interpretation, we examined both the 10 correct and 10 incorrect characters with the highest confidence according to the model. We initially assumed that the gender annotations found in the DraCor corpus were gold standard. However, through this examination we identified several annotation mistakes in the corpus. Among the 10 most confident `incorrect’ male predictions, were two female characters incorrectly annotated as male (\textit{Palas} and \textit{Glauca}). Likewise, among the 10 most confident `incorrect’ female annotations, we found an erroneously annotated character (\textit{Simón}). 

\paragraph{Attribution Model.}\label{attribution model results}
From the quantitative model evaluation, it is clear that the model is able to predict the gender of characters in Calderón's plays and even identify some annotation mistakes. However, it is not clear yet to what extent the model drew on features of the input that actually have to do with gender portrayal, despite our efforts to remove confounder features (see the discussion on lexical overfitting in \autoref{sec:methods}). By using an attribution model, we now analyze concretely on the basis of which tokens the model makes its classification decisions. In the following analysis, we list BETO tokens together with their integrated gradient scores, where negative scores indicate a tendency towards female speech, and positive scores a tendency towards male speech. Tokens with values closer to 0 are gender neutral, according to the model.

\begin{table}[tb!]
    \begin{tabular}{rc|rc}
        \hline
          Most masculine tokens & n &Most feminine tokens & n\\ \hline
        (``nosotras’’, 0.7763) & 2 &(``ton’’, -0.8746) & 1 \\
        (``\#\#jote’’, 0.6984) & 1 &(``cansada’’ -0.8000) & 1 \\
        (``Duque’’, 0.6460) & 1 &(``Ami’’, -0.6729) & 1 \\
        (``Emperador’’, 0.6105) & 2 &(``público’’, -0.6722)& 1 \\
        (``Majestad’’, 0.6073) & 3 &(``desesperada’’, -0.6178) &  2 \\
        (``escu’’, 0.6008) & 1 &(``contenta’’, -0.5764) & 1 \\
        (``tropas’’, 0.5860) & 1 &(``Todas’’, -0.4795) & 16 \\
        (``infantería’’, 0.5839) & 1 &(``esperaré’’, -0.4431) & 1 \\
        (``\#\#estre’’, 0.5824) & 1 &(``beso’’, -0.4076) & 1 \\
        (``encerrado’’, 0.5733) & 1 &(``minuto’’, -0.3882)& 1 \\
        \hline
        (``Alá’’, 0.5613) & 1 &(``amigas’’, -0.3841) & 2 \\
        (``Gobernador’’, 0.5448) & 2 &(``damas’’, -0.3609)& 4 \\
        (``Nevada’’, 0.5178) & 1 &(``\#\#ísimas’’, -0.3163)& 1 \\
        (``Fur’’, 0.5095) & 5 &(``Dis’’, -0.3154)& 2 \\
        (``camarada’’, 0.5076) & 2 &(``\#\#acas’’, -0.3066)& 1 \\
        (``alojamiento’’, 0.4962) & 3 & (``discursos’’, -0.2966) & 1 \\
        (``Fox’’, 0.4954) & 1 &(``\#\#món’’, -0.2954) & 1 \\
        (``soldado’’, 0.4922) & 2 & (``cuerpo’’, -0.2901 )& 1\\
        (``torneo’’, 0.4919) & 1 & (``comedia’’, -0.2895) & 1 \\
        (``criados’’, 0.4915) & 1 & (``celosa’’, -0.2844) & 1 \\ \hline
    \end{tabular}
    \caption{This table contains the 20 most polarized token attributions, as calculated by the character-level model.  $n$ is tokens frequency in the test corpus.}
    \label{tab:polarized_tokens}
\end{table}

In order to determine if there were clear themes in the tokens that the model judged as being more masculine or more feminine, we first examined the averaged attribution score of each token, calculated as the average attribution score of the token over all sentences in the validation set in which it occurs. \autoref{tab:polarized_tokens}  lists the 20 tokens deemed to be the most feminine and masculine, respectively.\footnote{For convenience, the Appendix provides a translation of this table into English as \autoref{tab:polarized_tokens_english}.} 

We  see several patterns in the attribution scores. Included in the most feminine tokens are several word endings or words with feminine endings (\textit{cansada}, \textit{desesperada}, \textit{contenta}, \textit{celosa}, \textit{todas}, \textit{\#\#ísimas}, \textit{\#\#acas}). These morphologically grounded indicators of female characters indicate that in Calderón's plays, female characters either refer to themselves or other women frequently. At the semantic level, we see that some of the most feminine words have some semantic connection to interpersonal relationships between characters (\textit{amigas}, \textit{beso}, \textit{discursos}, \textit{cuerpo}, \textit{celosa}). This indicates that, broadly throughout Calderón's works, female characters are more likely to discuss themes of love and relationships than male characters. Others among these words (\textit{cansada}, \textit{desesperada}, \textit{contenta}, \textit{celosa}) also demonstrate a depiction of women as emotional -- the women in these plays are on the one hand able to discuss their emotional inner worlds, but on the other hand depicted as driven or controlled by their emotions (\cite{drouet_2017}). 

This finding ties in with the results in \textcite{vsecla_2024}, who show that female characters across different languages and time periods are more likely to reference their emotions than male characters. Here we can see that Calderón's work somewhat conforms to these norms, but we will later see some evidence of the subversion of this trope in \autoref{sec:exp2-results}.

The male list contains several titles (\textit{Majestad}, \textit{Emperador}, \textit{Duque}, \textit{Gobernador}), which indicates the same pattern of more within-gender than across-gender interactions for male characters. Furthermore, these words indicate the importance of status in male-male interactions, arguably closely related to the topic of honor which was established as important across many works by Calderonian scholarship.\footnote{Note, however, the absence of words for honor itself (\textit{honor}, \textit{honria}), arguably since honor is relevant for both men and women, even though in different contexts.} Corresponding to the female indicators for conversations about relationships, the male list shows several words semantically related to combat (\textit{tropas}, \textit{infantería}, \textit{torneo}, \textit{soldado}), apparently the prototypical male topic. As previously mentioned, honor is of primary importance in Calderón's works. In addition to the fact that war is a typically male endeavor, in the \textit{siglo de oro}, male characters frequently use violence as a means to gain or restore honor (\cite{honig_1961}). It is therefore unsurprising that these themes feature prominently as `masculine’ words.

Curiously, \textit{nosotras} is the token most attributed as male, however, it is not spoken by a male character in the corpus. It appears in the validation set 8 times, and of the 8 characters who do speak this token, 7 of them are correctly classified as female characters. Upon further investigation, it was found that \textit{nosotras} appears in the training corpus, spoken by a female character that is mislabeled as male. Its listing as a female token is therefore the result of an annotation mistake, beyond the scope of our current study.

We also analyzed a list of top-40 male and female tokens with regard to their morphosyntactic properties, given that word classes or word types can also be used for stylistic gender portrayal (\cite{caliskan_2022}). We found that the most female tokens included significantly more word fragments (15), compared to the male tokens (10). Conversely, the male tokens included more nouns (24) compared to the female tokens (14). The female tokens also included slightly more adjectives (5) compared to male (3). Both used an equal number of verbs. In conclusion, there do not appear to be major differences at the level of word classes -- if anything, female characters tend to use less frequent words that tend to get split up into subword tokens.

Overall, this inspection of the features indicates that Calderón's gender portrayal -- within the space of features considered by the model -- has two main components: (a) at the interaction level, women and men both prefer interlocutors of the same gender; (b) at the content level, women are more concerned with the domestic sphere and men the military sphere. There do not seem to be major formal linguistic differences between the two genders.

Note that we only inspected the top tokens associated with either gender, and therefore this analysis cannot make any strong claim as to completeness. It is indeed likely that other topics arise from a further analysis of such tokens. Furthermore, the masking procedure we carried out during processing can be repeated after this type of analysis: If, for example, the analyst decides that tokens indicating patterns of interaction are not interesting in terms of gender portrayal, these tokens can be masked, and the model training repeated. The result is a model which should have the `next best’ set of features among its top-ranking gender cues.

\subsection{Experiment 2: Cross-Dressed Characters}
\label{sec:exp2-results}

\begin{table}
    \begin{tabular}{lllc}
        \hline
        Character & Gender & Play & Words Spoken \\ \hline
        Rosaura & Female & La vida es sueño & 3,479 \\
        Eugenia & Female & El José de las mujeres & 3,360 \\ 
        Lindabridis & Female & El castillo de Lindabridis & 2,842 \\
        Semíramis & Female & La hija del aire II & 4,997 \\
        Claridiana & Female & El castillo de Lindabridis & 3,332 \\
           \hline              
    \end{tabular}
    \caption{Cross-dressing Characters.} \label{tab:cross_dressers}
\end{table}

Cross-dressing characters are characters that adopt some traits of another gender during some scenes, dressing and speaking as a member of another gender. Such characters are a particularly interesting use case for our classifier-based gender portrayal method.

We obtained a list of five cross-dressing characters in four of Calderón's plays, listed in \autoref{tab:cross_dressers}, from the  \textit{Role Call} database.\footnote{See: \url{https://rolecall.eu/crossdressed\_characters/}.} Since we are not aware of other sources for instances of cross-dressing in Calderón's work, we treat this list as ground truth, although the possibility remains that the list is not exhaustive. All five characters are female characters cross-dressing as men: Rosaura, Eugenia, Semíramis, Claridiana, and Lindabridis. Each of these female characters subvert traditional roles by embodying manly traits such as intellectual aspirations, interest in war and hunting, physical strength, and resistance to marriage (\cite{araico_2017}). Information about in which scenes the characters are cross-dressing are taken both from the Role Call database and from looking at stage directions. Unfortunately, there are many inconsistencies, and the cross-dressing is not always explicitly stated in the stage directions. In the cases where the stage direction indicates cross-dressing, we use this information, otherwise we revert to the Role Call list.

For the ecological validity of this experiment, we ensure that all cross-dressing characters are part of the validation set -- that is, the gender classification model is not trained on any cross-dressing characters. As motivated above, we use our trained model that aggregates predictions at the scene level. We do not carry out any specific prediction for these characters, but simply analyze the predictions of the model, as in Experiment 1, but now at the level of specific characters. We structure our analysis along two hypotheses.

\begin{figure}
  \centering
\includegraphics[width = 1\textwidth]{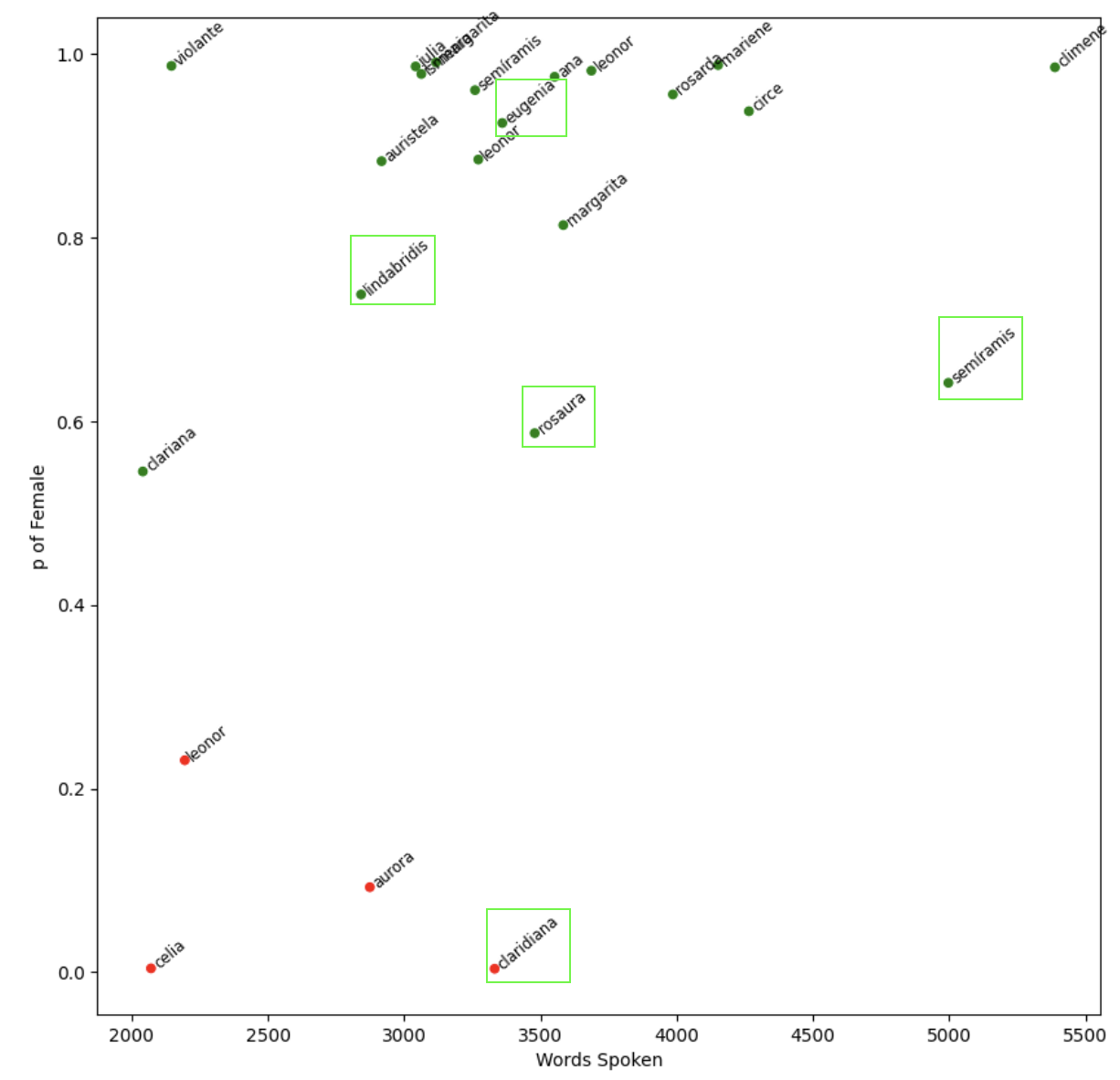}
  \caption{Experiment 2: Probability that the character is predicted as female (prediction made at our first input level, all lines) for all female characters speaking more than 2000 words (green dot = predicted as female, red dot = predicted as male, green box = cross-dressing). Note: The highlighted instance of Semíramis is from la \textit{Hija del Aire II}.}
  \label{fig:cross-dressing-results}
\end{figure} 

\paragraph{Hypothesis 1: Lower Confidence for Cross-Dressers.} If the model picks up on gender differences used strategically by Calderón, the predictions for the cross-dressing characters should be less confident about their gender than for other characters. \autoref{fig:cross-dressing-results} shows the model's overall confidence at the character level for all female characters from the four relevant dramas (\autoref{tab:cross_dressers}), with the five cross-dressers marked by green boxes. Compared to other female characters that speak the same number of words, the model predicted cross-dressing characters to be female with less confidence (mean = 0.78 compared to mean = 0.89). Only four female characters are predicted with comparably low confidence (Clariana, Leonor, Celia, and Aurora), all speaking comparatively little. This pattern makes sense, given that cross-dressing characters mix female-like and male-like speech, providing conflicting evidence to the model.

Additionally, the model struggles with the overall classification of the cross-dressed characters more than with  non-cross dressing ones (the accuracy is 4/5=0.80 compared to 19/23 =0.83). Again, this is expected, given that we would expect a data-driven model to classify a cross-dressing character either way, depending on the extent of their cross-dressing. Indeed, the `misclassified’ character is Claridiana, who cross-dresses in the whole play (see below).

\begin{figure}[tb!]
  \centering
  \includegraphics[width=0.75\linewidth]{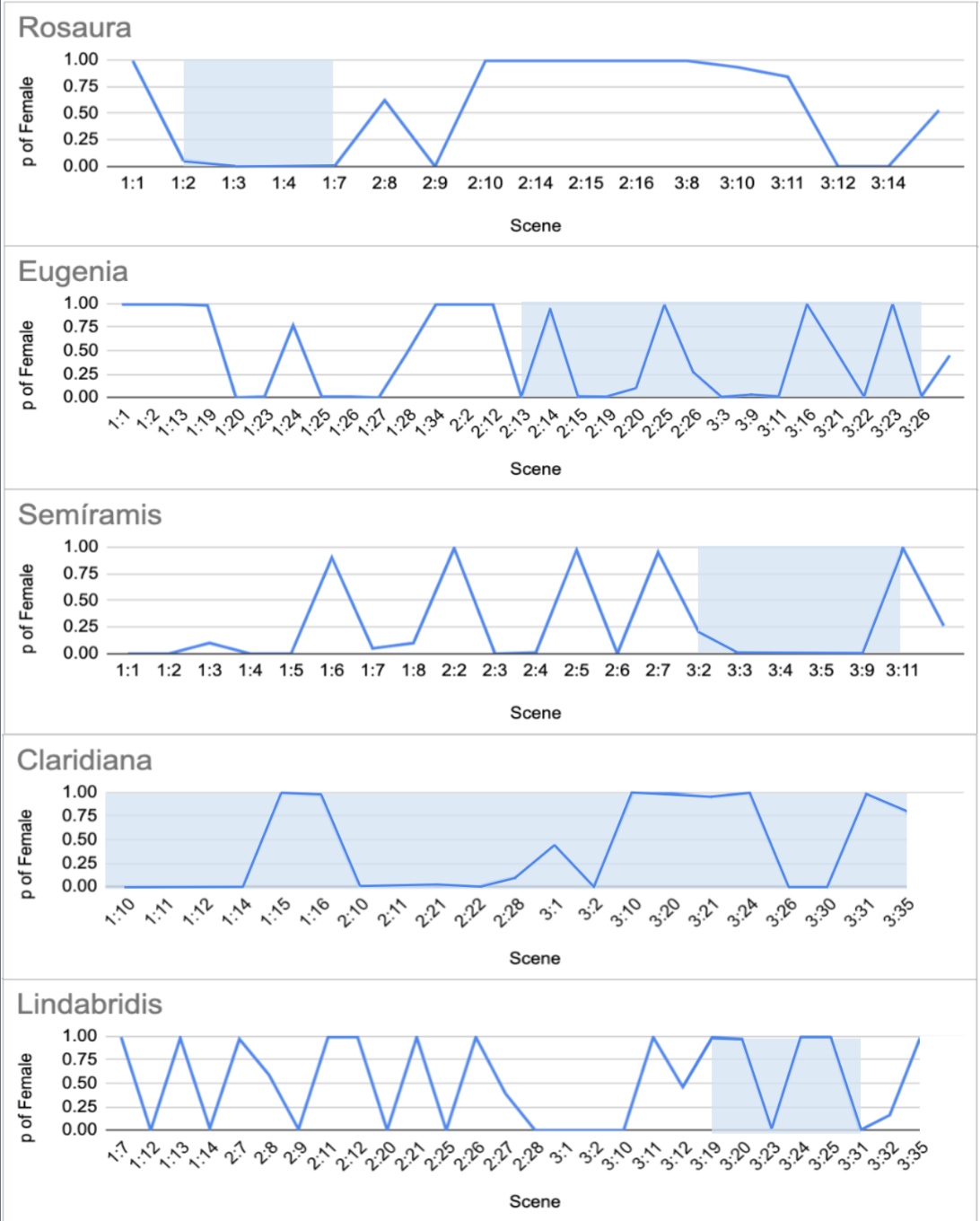}
  \caption{Experiment 2: Analysis of cross-dressing characters by scene. Cross-dressing scenes are indicated by blue bars. y axis represents the probability that the character will be predicted as female.}
  \label{fig:crossdressers-by-scene}
\end{figure}

\paragraph{Hypothesis 2: Scene-By-Scene Predictions.} If the model does a good job of picking up gender cues in a fine-grained manner, it should be able to tell in which scene a character is cross-dressed. We test this hypothesis on the four relevant dramas, and \autoref{fig:crossdressers-by-scene} shows the results: The cross-dressing scenes (according to the Role Call database, and in some cases identified by us in the few instances where stage directions indicate cross-dressing) are shown by a blue shading, and the model predictions are indicated on the y axis: the probability that a character was predicted to be female. Along the x axis are all the scenes in which a character speaks (ex. 1:2 on the axis represents Act 1, Scene 2). We find that, with a proportion higher than chance, cross-dressing characters were predicted as male for the scenes in which they were cross-dressing -- the model is able to pick up on instances in which these female characters are behaving in a stereotypically masculine fashion. 

The drama \textit{La hija del aire II} focuses heavily on themes of governance and war as well as the mythical figure Semíramis, the queen of Asiria. Semíramis is described as being competent, a brave warlord, cruel and treacherous, but also evoking feminine sensuality. In the act three, after the death of her son, Semíramis disguises herself as her son in order to assume his place on the throne. In all of the scenes that Semíramis dresses as her son, the model predicts her as male, whereas through the rest of the play the model mostly predicts her as female.

In the story of \textit{El Castillo de Lindabridis}, Lindabridis is a princess who travels the world in a flying castle looking for a husband. In this work, Claridiana is a female character who disguises as a male throughout the play in order to win the hand of Lindabridis. Claridiana engages in the trials of courage and combat required to marry Lindabridis. Claridiana is predicted by the model to be male in 14 out of 21 scenes in which she occurs. Lindabridis also disguises herself as a man in the third act of the play and in 6 out of 11 scenes of the third act she is predicted as male. 

Finally, the model does a good job in \textit{La vida es sueño}, in which Rosaura disguises herself as a man in the first act of the play in an attempt to seek revenge on a man who took her honor. The cross-dressing is clearly picked up by the model because she is predicted as male in the majority of the first act (4/5 scenes) and mostly as female in the rest of the acts (8/11 scenes).

From these cases, we can see that not only is the the model able to predict the gender of characters with high accuracy, it is also able to detect specific scenes where female characters are cross-dressing. Therefore, we can see that the social context of the characters has a large influence on the predictions of the model. Still, a more fine-grained analysis of the individual plays would help contextualize the model's mispredictions at the scene level.

In any case, the model is still able to predict many of these cross-dressing characters as women, but with lower confidence, indicating the complexity of the characters -- they are women and embody traditional female traits, while using cross-dressing as a means to transcend their social role as women and accomplish tasks in the public social sphere, wield power, and take agency of their own stories (\cite{woods_2013}).

\section{Discussion and Conclusions}

In this paper, we have taken a first step towards a `scalable reading’ investigation of gender depiction on a large corpus of \textit{comedias} written by Calderón de la Barca. We make our implementation publicly available.\footnote{Software can be found here: \url{https://github.com/allisonakeith/calderon-gender-prediction} (DOI: \url{https://doi.org/10.5281/zenodo.15639883}).} We have proposed an analysis procedure where we train a gender classifier on the text (or parts of it) spoken by each character as an aggregation method. We then decompose the classifier's predictions using an attribution method to critically assess the features that the classifier bases its predictions on. Since many attribution methods, including the integrated gradients we use here, are provably correct, this procedure can avoid the use of heuristic methods to identify keywords or topics, such as measures of association or keyness (\cite{schroeter_2021, vsecla_2024}) or topic modeling (\cite{wallach_2009}). Our method, nevertheless, keeps the domain expert at center stage and lets them determine the interpretation of the predictions and attributions presented.

The contributions for computational linguistics lie at the methodological level, in better understanding what properties of the setup are important for the success of the method. We investigate the best level at which to present the speech to the classifier and find that presenting a person's complete speech at once leads to robust results, but that classifying individual sentences and aggregating the results by probability does not only yield an improvement in overall accuracy, but also makes the predictions more accurate for the smaller gender category (female). Despite their hunger for data, the models also perform reasonably well when asked to classify character speech from single scenes, allowing us to trace the plot in dramas involving cross-dressing characters.

The contribution for literary studies lie at the substantive level, using the method to gain a better understanding of gender portrayal in Calderón's \textit{comedias}. We have shown that male and female characters broadly demonstrate adherence to gendered speech in a way that is classifiable at the sentence level. Our findings bolster prior observations by \textcite{vsecla_2024}, finding similar thematic differences between male and female characters (female characters making reference to their internal worlds, for example). At the same time, our qualitative analysis on the example of Rosaura shows that, at the character level, Calderón's gender representation can be non-monolithic and complex (\cite{arellano_2015}) -- and our computational tools provide us with a way of quantifying this intuition. Overall, in this way, our models produce findings that align with the existing hermeneutical studies, but can extend them to many of Calderón's lesser known \textit{comedias}. Concretely, we find that the main cues of gender identity that our models pick up regard modes of interaction (within-gender vs. cross-gender) as well as spheres of existence (the domestic and personal for female characters vs. the military and political for male characters). 

Another surprising but positive finding in our study was that several findings that looked wrong or at least counterintuitive at first glance turned out to be due to annotation errors in the corpus or the data bases we used, strengthening our belief in the accuracy of the method's output and indicating its usefulness to uncover such errors in the annotation, an application known in computational linguistics as ``error mining’’ (\cite{dickinson_2015}).\footnote{We will correct the annotation errors we found and will submit these changes for publication with CalDraCor.}

More generally, our work indicates that Calderón's dramas -- at least in modernized and normalized versions -- do not provide major obstacles to the use of current language model-based NLP, and thus lays the ground work for future use of scalable reading approaches in Calderonian studies.

In terms of limitations, the study reported here was based on a gender classifier with a BERT architecture, which comes with a set of constraints. These include the use of a model-specific tokenizer which cannot be easily adapted to the specific corpus and  which may make some of the attributions difficult to understand, as well as the 512-token limit of the classifier. Most seriously, we limited ourselves to only the text spoken by each character, not taking into account dramatic structure or interpersonal relationships. Due to this simplification, the model is limited to lexical cues, and could not identify gender-based patterns like (hypothetically) `men always speak first’. While our study shares this limitation with much work in stylometry (\cite{evert_2017}), it is possible to present such structural information to classifiers as well (\cite{kim_2019}), in particular in more recent models with longer context windows (\cite{zaheer_2020}). We leave such endeavors to future work.

We conclude by sketching three more avenues for future research. The first one is to move beyond gender to other character traits such as socio-economic status. Honor is a key theme in the works of Calderón, and it is intimately tied to socio-economic status and hierarchy (\cite{greer_2017,ibarreche_2020}). Particularly, noble characters spend many plays discussing the losing and gaining of honor. Characters of high socio-economic status are more likely to be well educated and thus speak differently to lower class character or servants (\cite{woods_2013}). \textit{Siglo de oro} authors also utilize character archetypes which are tied to different socio-economic statuses such as the \textit{gracioso} -- a servant character who serves as comic relief. These are frequently labeled in the cast sheet, however not all instances of this character type are annotated in our corpus. Therefore, it may be possible to utilize similar classification methods to identify socio-economic status and enrich the annotations in CalDraCor. Identifying archetypes may require the use of additional processing tools, such as clustering. This information can then be used to identify structural patters, in Calderón's work.

A second direction could be the use of our approach to find structure in the large number of works by Calderón, looking for example at differences in gender portrayal between his tragedies and his comedies, or conversely using gender depiction as evidence to argue for or against certain genre delineations (\cite{lehmann_2023}). Another possible question to ask is whether Calderón's gender portrayal remained constant over the long period of his productive career, or whether it changed.

Finally, future work could apply the  methods developed here to other dramas by \textit{siglo de oro} authors such as Lope de Vega. This would permit us to understand the extent to which gender portrayal we observe in Calderón's \textit{comedias} is a general feature of the period or author specific. This would require overcoming practical hurdles, though, since other authors' work are generally not as easily accessible and consistently represented as Calderón's in CalDraCor.

\paragraph{Acknowledgements.} This work was conducted as part of the 'Identifying Regularities in the works of Pedro Calderón de la Barca' project (508056339) funded by the DFG Priority Programme / Schwerpunktprogramm 'Computational Literary Studies' (SPP 2207).

\printbibliography

\section{Appendix}

\subsection{Model training details}
\label{sec:training-details}

In order to use the model for our gender classification task, we add a binary classification head on top. We use the character lines from CalDraCor together with the ground truth gender labels as data to tune this classification head. We used the Adam optimizer, a standard gradient descent based optimizer \parencite{zhang_2018}, and considered a range of reasonable hyperparameter values recommended in the literature.  The best values for each models are listed in Table \ref{tab:training-details} This resulted in a learning rate of 1e-05 for each model. We implemented early stopping to prevent overfitting of the training data.

\begin{table}[h]
    \centering
    \begin{tabular}{lccc}
    \hline
Input Level & Batch Size & Number of Epochs & Learning Rate \\ \hline
Character & 24 & 5 & 1e-05  \\
Scene & 32 & 7 & 1e-05 \\
Utterance & 32 & 14 & 1e-05  \\ \hline
    \end{tabular}
    \caption{Details on Model training. All models use BETO as the base for creating word embeddings, and are fine-tuned with our data on binary classification. Best performance was found by testing model with various parameters, and implementing early stopping so that model stops at the best performing epoch}
    \label{tab:training-details}
\end{table}

\subsection{Indicative tokens: English translation}

We include here for reference English translations of the words with most extreme attribution scores (-1 to 1) from the character level test data. We also include the number of times each word occurs in the test data. In cases where word fragments appeared (usually they just appeared once in the test data) we identified the words to which the word fragments belong. In the case of 'ton' the word fragment occurs before the length cut off and therefore the model does not consider, and 'ton' alone doesn't have semantic meaning, so no translation is given. 

\begin{table}
    \centering
    \begin{tabular}{rc|rc}
        \hline
          Masculine Attributions & n &Feminine Attributions & n\\ \hline
        ('we (fem.)', 0.7763) & 2 &('ton\#\#', -0.8746) & 1 \\
        ('\#\#jote<- from Don Quijote', 0.6984) & 1 &('tired', -0.8000) & 1 \\
        ('Duke', 0.6460) & 1 &('Ami\#\# <- friend (fem)', -0.6729) & 1 \\
        ('Emperor', 0.6105) & 2 &('public, audience', -0.6722)& 1 \\
        ('Majesty', 0.6073) & 3 &('desperate(fem.), hopeless', -0.6178) 2 &\\
        ('escu\#\#' <- shield, 0.6008) & 1 &('contenta', -0.5764) & 1 \\
        ('troops', 0.5860) & 1 &('all (fem.)', -0.4795) & 16 \\
        ('infantry', 0.5839) & 1 &('will wait', -0.4431) & 1 \\
        ('\#\#estre'<- master, 0.5824) & 1 &('kiss', -0.4076) & 1 \\
        ('encerrado', 0.5733) & 1 &('minute', -0.3882)& 1 \\
         \hline
        ('Allah', 0.5613) & 1 &('friends (fem.)', -0.3841) & 2 \\
        ('Governor', 0.5448) & 2 &('lady', -0.3609)& 4 \\
        ('Nevada', 0.5178) & 1 &('\#\#ísimas'<-most beautiful, -0.3163)& 1 \\
        ('Fur\#\#'<-wrath, 0.5095) & 5 &('Dis\#\#' <- pretend, conceal -0.3154)& 1 \\
        ('comrade', 0.5076) & 2 &('\#\#acas', -0.3066)& 1 \\
        ('camp, accommodations, 0.4962) & 3 & ('discourses, speeches', -0.2966) & 1 \\
        ('Fox', 0.4954) & 1 &('\#\#món' <- sermon, -0.2954) & 1 \\
        ('soldier', 0.4922) & 2 & ('body', -0.2901 )& 1\\
        ('tournament', 0.4919) & 1 & ('comedy', -0.2895) & 1 \\
        ('servants, foot men', 0.4915) & 1 & ('jealous (fem.)', -0.2844) & 1 \\ \hline
    \end{tabular}
    \caption{English translations of the most masculine and feminine words in the test corpus. In cases where gender is indicated by the word ending, this is noted. Some word segments have no translatable meanings. In these cases, }
    \label{tab:polarized_tokens_english}
\end{table}

\end{document}